\documentclass[10pt,twocolumn,letterpaper]{article}

\usepackage{ijcb}
\usepackage{times}
\usepackage{epsfig}
\usepackage{graphicx}
\usepackage{amsmath}
\usepackage{amssymb}
\usepackage{booktabs}
\usepackage{url}
\usepackage{authblk}
\usepackage{comment}

\ijcbfinalcopy 

\ifijcbfinal\fi

\begin{document}

\title{Long-Range Biometric Identification in Real World Scenarios: A Comprehensive Evaluation Framework Based on Missions}







\author{Deniz Aykac, Joel Brogan, Nell Barber, Ryan Shivers, Bob Zhang,  Dallas Sacca, Ryan Tipton,
Gavin Jager, Austin Garret, Matthew Love, Jim Goddard, David Cornett III and David S. Bolme\\

Oak Ridge National Laboratory \\
1 Bethel Valley Road, Oak Ridge, TN 37831, USA\\
\texttt{aykacdb@ornl.gov}
}

\maketitle
\thispagestyle{empty}

\begin{abstract}
The considerable body of data available for evaluating biometric recognition systems in Research and Development (R\&D) environments has contributed to the increasingly common problem of target performance mismatch. Biometric algorithms are frequently tested against data that may not reflect the real world applications they target. From a Testing and Evaluation (T\&E) standpoint, this domain mismatch causes difficulty assessing when improvements in State-of-the-Art (SOTA) research actually translate to improved applied outcomes. This problem can be addressed with thoughtful preparation of data and experimental methods to reflect specific use-cases and scenarios.

To that end, this paper evaluates research solutions for identifying individuals at ranges and altitudes, which could support various application areas such as counterterrorism, protection of critical infrastructure facilities, military force protection, and border security. We address challenges including image quality issues and reliance on face recognition as the sole biometric modality. By fusing face and body features, we propose developing robust biometric systems for effective long-range identification from both the ground and steep pitch angles. Preliminary results show promising progress in whole-body recognition. This paper presents these early findings and discusses potential future directions for advancing long-range biometric identification systems based on mission-driven metrics.

\end{abstract}


\section{Introduction}
\label{sec:introduction}

The increasing complexity of modern security scenarios necessitates the development of advanced biometric identification systems that can operate effectively over a wide range of distances, altitudes, and environmental conditions. The Intelligence Community (IC) requires such technology for various critical applications, including counterterrorism, protection of vital infrastructure facilities, military force protection, and border security. However, effective implementations of current biometric systems are hindered by several major challenges.

\begin{figure}
    \centering
    \includegraphics[width=0.95\linewidth]{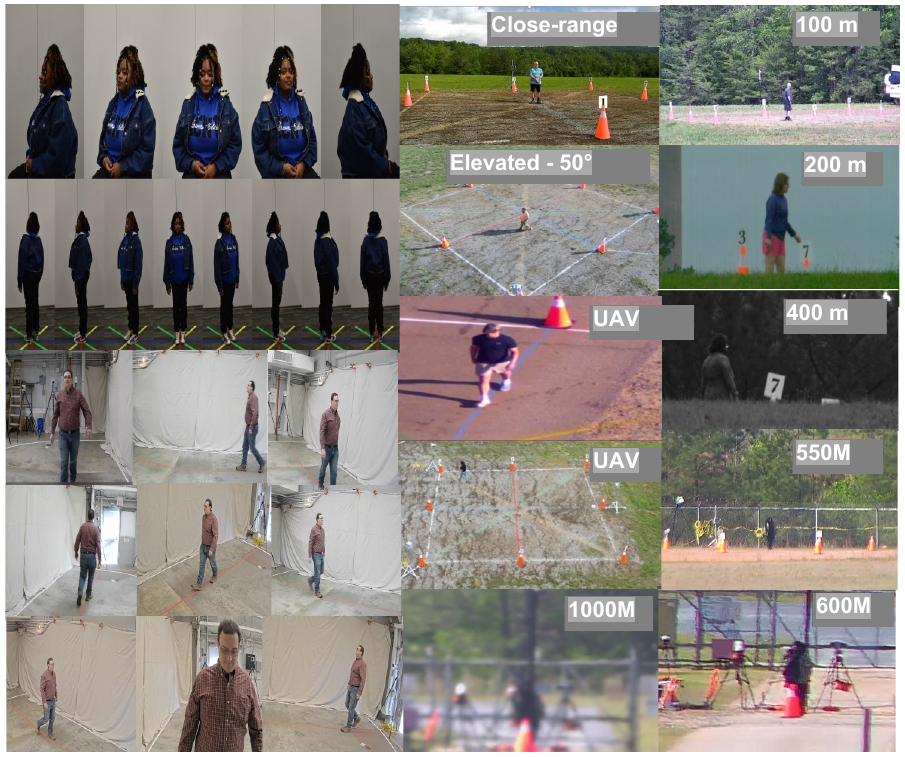}
    \caption{
Example dataset showcases a collage with two distinct sections. The left portion contains gallery images and video frames sourced from a controlled collection. The right section comprises various video frame probes.}
    \label{fig:samples}
\end{figure}

Firstly, image quality issues arising from factors such as motion blur, atmospheric turbulence, and resolution limitations significantly degrade system performance. Secondly, most algorithms have been developed based on low-pitch angle views of people, which may not be effective in scenarios characterized by significant pitch angles, such as those encountered when using UAVs or cameras mounted on buildings. Lastly, the over-reliance on face recognition as the primary biometric modality limits system efficiency and accuracy, especially in situations where faces are not visible or when other body features provide more reliable identification cues.


To support research that addresses these challenges, this paper evaluates performance by current research solutions on face, body, and gait imagery targeting operationally relevant mission areas. These mission areas include close- and long-range face recognition, long-range recognition using multiple fused modalities, recognition from UAV platforms and elevated cameras, atmospheric turbulence mitigation, gait recognition, and identification without any full views of the face. Each mission area introduces a unique set of experimental problems that algorithms must handle to ensure robust system performance and accuracy.


We select video recorded from ranges up to 500m, pitch angles up to 30 degrees and evaluate five software systems specifically designed to process short video clips and perform biometric matching on whatever features are available. Unlike face recognition systems, these are built to handle complex input comprising views of the person from any angle with potential occlusions while still addressing quality issues inherent to long-range and elevated imagery.

The solutions evaluated in this paper were developed by independent research teams whose goal is to develop an advanced biometric identification system that can operate effectively under a wide range of conditions to support the IC's critical mission requirements.

In the subsequent sections, we provide a comprehensive evaluation framework based on missions. We begin with discussing related work on biometric recognition under challenging conditions. Then, we describe our methods of data collection and preparation for this evaluation, provide an overview of the unified application programming interface (API) we use to evaluate algorithm performance, and detail the mission areas selected for this work. Finally, we explain our analysis design, discuss results, and lay out a path forward to continue research as algorithms continue to mature.
\section{Related Work}
\label{sec:related-work}

The limitations of accurate human identification at stand-off distances has been of interest of both the IC and research community for years. Earlier works such as \cite{yao2008improving} acknowledge that a major limiting factor is the lack of available datasets on which to train and evaluate new biometric models. Authors in \cite{li2016toward} and \cite{neves2016biometric} demonstrate the need for systems in low-resolution and unconstrained collection environments. Research such as \cite{luevano2021study}, \cite{schlett2022face}, \cite{meng2021magface}, and \cite{terhorst2023pixel} explore the impact that the number of pixels on the face as well as the quality of the face video itself have on facial recognition systems, which has spurred entities such as IARPA and DARPA to invest in datasets that can help improve biometric performance at long ranges and lower qualities~\cite{cornett2023expanding,calhoun2016darpa} The utility for facial recognition systems that can perform accurately at long distances is also addressed in \cite{bolmerifle} and \cite{park2013face}.

In recent years, the prevalence of high-quality, lower cost UAVs has lead to their widespread usage for various applications. Those systems with mid-to-high quality imaging sensors lend themselves especially to use cases involving human detection, identification and tracking. Research in \cite{aruna2020p}, \cite{balint2018uavs}, \cite{sarath2019unmanned}, \cite{shanthi2023smart}, 
and \cite{rostami2022deep} investigate pairing facial detection and recognition systems with these types of UAVs. Research as seen in \cite{diez2023dynamic} even acknowledges the limitations of existing biometrics systems to address the challenges of uncontrolled lighting, distance and pitches demonstrate the need for new algorithms to overcome these challenges.

The current related body of research coupled with the growing needs of the IC demonstrate the need for the advances detailed in this paper.

\section{Data Preparation} 
\label{sec:data-preparation}

This section lays out the three components that comprise our methods to evaluate state-of-the-art solutions for identifying individuals at ranges and altitudes. The Data Collection section details our endeavor to produce a high-quality, statistically relevant dataset to provide both training and testing data for biometric recognition systems. The Data Curation and Partioning section provides insights into how the data is structured and organized to evaluate the various mission categories. Finally, the Common API section summarizes the custom API used across all solutions so that a fair and non-biased evaluation can be performed.
\setlength{\textfloatsep}{5pt }

\subsection{Data Collection}

It is necessary to build a unique biometric image and video dataset to evaluate algorithm performance. This dataset consists of planned biometric collection events held in different parts of the United States during different seasons, ultimately resulting in diversity of subject demographics, terrain, and weather conditions. The dataset includes images and video of subjects in a variety of controlled and uncontrolled situations using a variety of sensors. 

Subjects participated in the experiments only for one day and were required to wear different clothing sets. In the field, for example, the probes had clothing set 1 and gallery had clothing set 2 for the same subject.  This ensures that algorithms are performing recognition based on persistent biometric signals from the face, body, and patterns of movement and not simply recognizing their clothing.

The indoor phase consists of face and whole-body images captured from multiple pitch angles up to 50° and the full rotation of yaw angles at 45° increments. This phase also includes video captured by COTS sensors placed in a semicircle around the collection area to record subjects walking and using their cell phones from multiple angles. Subjects repeat the indoor collection process in two different sets of clothing.  

During the outdoor data collection, subjects are recorded while they are standing still facing different directions, walking along prescribed paths, and performing a random walk. Conditions and types of data available may vary considerably based on characteristics of each collection site. Outdoor data is collected without regard for the prevailing weather conditions. Data is collected using close-range video cameras at various pitch angles from sensors loaded on as many as three different UAVs and from several long-range sensors set at a range of distances from 100m to 1,000m. The long-range sensors comprise a selection of COTS and industrial USB cameras equipped with specialized optics. At most distances, there is a camera set to capture the whole-body (WB camera) and a second camera set to capture a closer view of the face, which may not include the entire body (Face camera). The media samples from both indoor and outdoor collection are shown in Fig.~\ref{fig:samples}.

\subsection{Data Curation and Partitioning}
In this section, we present the steps to curate the data from all the sensor measurements and to formalize the probe and gallery compositions for evaluation.

\textit{Step 1}: During data collection, researchers use a custom desktop application to record timestamps for when each subject begins and completes a recorded activity. In the first step of post-processing, individual images and video segments are associated with a particular subject based on these timestamps. Longer videos containing footage of multiple subjects are cut along these timestamps so that the resulting dataset is comprised of videos of one subject performing one activity in one clothing set. This is made possible via precise time synchronization of all sensors. The length of video segments varies depending on the activity being performed and, in the case of some activities, the subject’s mobility and natural walking pace.

\textit{Step 2}: Additional data is recorded alongside videos during the data collection which describe weather conditions, atmospheric conditions, subject demographics, sensor positioning, and sensor hardware. This information is compiled such that each video segment can be associated with the corresponding metadata. Weather and atmospheric metadata that is tracked includes: temperature, wind chill, heat index, relative humidity, wind speed, wind direction, barometric pressure, and solar loading. This data, along with turbulent fluctuation recordings measured by a scintillometer, are recorded every minute during the data collection. Video segments are associated with the weather and atmospheric data aligned with the minute of its starting timestamp. Detailed sensor information such as minimum / maximum focal length, serial number, model number, manufacturer name, and sensor location are also included in the metadata describing each video segment. 

\textit{Step 3}: Once the data has been organized as described in Steps 1 and 2, it is partitioned into training and test sets. This is done in a manner designed to create consistency of demographic distributions between the two sets. Every subject is assigned to either the training set or the test set, and the data is organized into its final directory structure. An XML file is generated for each piece of media to store the metadata described in Step 2 and the annotations described in Steps 4 and 5.  A schema definition is used to validate XML files after metadata has been added to detect improperly formatted or corrupted data.

\textit{Step 4}: Automated annotations are generated using a chain of open-source and pretrained models. WB detection is done with YOLOv5 and a fine-tuned version is used on long-range and aerial videos \cite{yolov5}. 3D human mesh reconstruction with Meshtransformer and 2D keypoint estimation with DARK is performed on the WB detection results \cite{Zhang_2020_CVPR, DBLP:journals/corr/abs-2012-09760}. Re-ID with DG-Net++ is then performed on the pose results to determine whether or not a WB detection is the intended subject or not \cite{zou2020joint}. The pose information helps narrow the gallery to reference images at a similar yaw angle to the detections. BoT-SORT is used for track generation, which leverages the Re-ID results for better track consistency \cite{aharon2022botsort}. Finally, various post-processing steps are performed such as estimating the head bounding box from the 3D mesh and 2D keypoints.

\textit{Step 5}: Select video frames are sent out for manual annotation either for validation or correction of suspected automated annotation errors. Mainly, these consist of a couple frames per track to verify the correct subject was denoted, a non-subject made it into the video, or the detection was a non-person. These results are then merged with the automated annotations. The metadata information such as type, method of collection along with examples are summarized in Table ~\ref{tab:metadata}.

\textit{Step 6}: The test set is further partitioned into gallery and probe sets for evaluation. Gallery media is used to build a database of known identities and probe media is imagery of subjects whose identities are presumed unknown. Probe imagery is compared to gallery enrollments and the results are used to compute the metrics we use to evaluate system performance. The evaluation gallery sets contain data for distractor subjects who are not part of the probes to simulate a larger gallery. We perform additional data partitioning for each specific mission area. These more granular designations are described in detail in Section ~\ref{sec:mission-areas}. Balancing the subjects for mission categories was applied during the probe selection process but variation in the number of subjects and the samples per mission still exist as shown in Table ~\ref{tab:mission}. 

\textit{Step 7}: The probe set for mission analysis is organized into two major partitions: Face Restricted and Face Included. In the Face Restricted probe set, all faces are either occluded, low resolution (\(<\)20 pixel head height), or not visible. The Face Included probe set contains data in which the face is visible and the head height is at least 20 pixels. The expectation is that there is at least one detectable face for every subject in every probe. Since width can vary more significantly with the changes of the yaw angle of the head, height is chosen as the basis for face detection. To establish whether the subject is facing the camera at any point during the 5-15 second video segment constituting a probe, we estimate the subject's yaw angle relative to the camera. A subject is considered to be facing the camera if the yaw angle is estimated to fall between \(-110^{\circ}\)  and \(110^{\circ}\). Two different techniques are utilized to achieve pose estimation: \textbf{Automatic:} The head portion of the 3D human mesh reconstruction is used to derive the yaw, pitch, and roll for person detected. Specifically, the Kabsch-Umeyama algorithm gives a rotation matrix from a reference 0 yaw, pitch, and roll head \cite{kabsch1978discussion, umeyama1991least}. \textbf{Activity timing:} Using RealSense cameras and elevated views, it is possible to track the subject's position within the collection box. This identifies the different segments of the standing and structured walking activities. Directions during these segments are well known and can be used to label the expected yaw angle relative to each camera position with accuracy.
 
\textit{Step 8}: Face Included and Face Restricted partitions are further split into treatment and control sets for evaluation. Control set consists of cameras positioned on the ground and at close range of less than 75m. For some evaluations, these cameras serve to set expectations for traditional performance. Most of these close-range cameras will have a wide field of view to capture the entire walking area and therefore may be at the low end of resolution needed for accurate face recognition. The treatment set consists of cameras from long-range, elevated, and UAV platforms. The treatment probes are given more weight in the probe selection process compared to the control probes as they are the focus of the project.


\begin{table*}
    \caption{Types of metadata collected}
\begin{center}
\small
\begin{tabular}{|c|c|c|}
\hline
Metadata Type & Method of Collection & Examples\\
\hline
\hline
Weather & At collection & Temp, humidity, wind speed\\
\hline
Demographics & At collection & Age, gender, height\\
\hline
Timing & At collection & Video start, end time\\
\hline
Camera & At collection & Make, model, resolution, focal length\\
\hline
UAV & At collection & Alt., GPS, camera telemetry\\
\hline
Face Bounding Boxes & Video processing & Location, size, tracks\\
\hline
Whole Body Bounding Boxes & Video processing & Location, size, tracks\\
\hline
Subject Identity & Annotation & Confirmation of identity\\
\hline
Additional Bounding Boxes & Annotation & Other objects and bounding boxes\\
\hline
Occlusion Information & Annotation & Info on occlusion of face/body\\
\hline
Clothing Information & Annotation & Type of clothing, shoes, accessories \\
\hline
\end{tabular}
\end{center}
\label{tab:metadata}
\end{table*}

\section{Common API}
\begin{figure}
    \centering
    \includegraphics[width=0.9\linewidth]{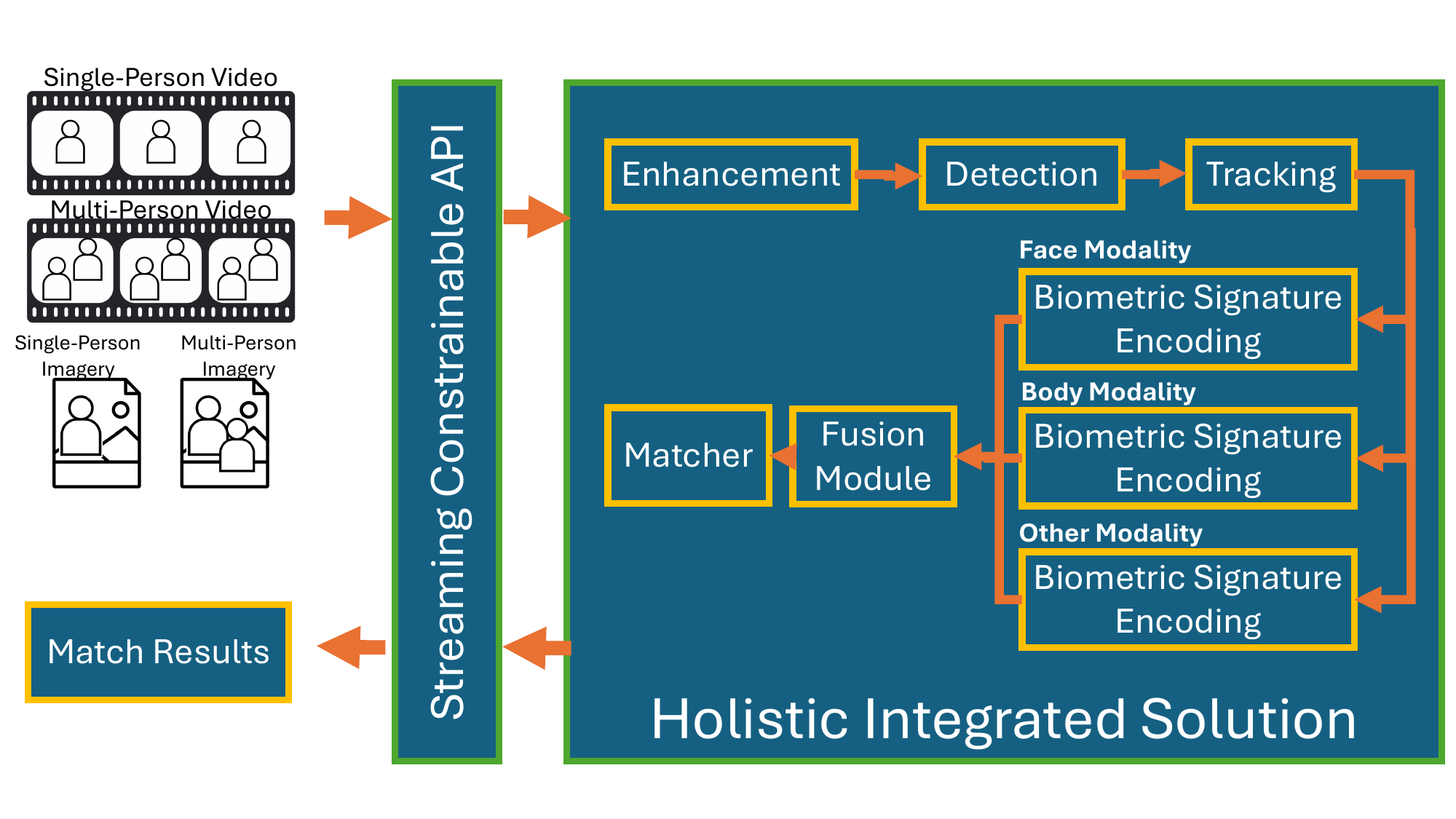}
    \caption{Diagram of an Integrated Holistic System (HS).}
    \label{fig:integrated_system}
\end{figure}

The API leveraged in this work took inspiration from previous works in \cite{gunther2012open}, \cite{golec2020biosec}, \cite{bolme2020face}, and \cite{brogan2023vdisc,brogan2022ornl}. It was built for this evaluation program and was guided by the same principles of maintaining operational relevance.  To that end, the API has 3 main purposes: 1) To provide an algorithm harness with a front-end common set of callable functions that can abstract away complex back-end SOTA biometric algorithms, 2) to enforce strict standards on research code that help guarantee stable and predictable execution patterns, and 3) to operate as a data-passing middle-man, which allows the API to simulate the types of data and usage behaviors that would reflect those of a deployed system. Figure~\ref{fig:integrated_system} shows a general overview of the API that is meant to interface with both streaming data and SOTA integrated biometric solutions to provide a realistic operational scenario. These end-to-end biometric solutions incorporate many internal algorithms across multiple biometric modalities, along with database-level storage and caching capabilities, to ultimately produce a single set of match scores for verification and search requests. For this reason, we refer to these SOTA solutions as Holistic integrated Solutions (HS). To evaluate an HS, the API ingests signature sets (sig-sets) similar to those introduced in~\cite{flanagan2011ice} and used in numerous biometric challenges~\cite{scheirer2016report}, which define and provide locations to subsets of video and imagery pertaining to the missions outlined in Section~\ref{sec:mission-areas}. The API then iterates over these sig-set-defined media subsets, loads the media, and subsequently streams the media to the HS.  Because the API is constrainable, it can be configured to strip out, modify, or reformat media and its metadata to better simulate real-world operating scenarios.  For instance, some scenarios may allow for metadata that provides camera information such as make, model, frame rate, operating conditions, etc., while other scenarios may need to simulate that metadata being unavailable to an HS at ingestion time.


\section{Mission Areas}
\label{sec:mission-areas}

In this section, we explain the mission areas selected for this analysis and provide examples of the kinds of use cases that may be captured by each of them. Table ~\ref{tab:mission} provides a brief overview of the mission areas discussed in this paper. Based on use case, some applications may focus on face recognition performance, whereas others might require the fusion of multiple modalities. The modality may also be determined by the biometrics available in galleries and watchlists. Currently, the barrier to entry for face recognition is relatively low due to the high availability of both frontal face imagery and highly accurate software systems. Other modalities such as body and gait are as yet less sophisticated and require significant performance advancement before they will be widely adopted. There is also a much smaller corpus of ground truth data for these modalities. Evaluating performance for these specific mission areas may also inform areas in which additional research is needed to achieve suitable accuracy for deployment in a real world application.

\textbf{Experimental Control}: We use imagery captured by high resolution surveillance cameras positioned at eye-level. This relatively "easy" data is used to establish a performance baseline.

\textbf{Close Range Face}: This area focuses on imagery with high resolution, frontal profile views of the face, simulating environments like building entrances, choke points, and security checkpoints. Typically, these areas are well-suited to face recognition because there is a high degree of certainty that people will pass through a contained space and face a certain direction. We use close range video from cameras positioned at eye-level or slightly elevated.

\textbf{Close Range Body}: This mission area is a super set of Close Range Face that includes all views of the face along with body and gait signatures. These videos are taken from high resolution surveillance cameras at close-range positioned at eye-level or slightly elevated. This mission can be used as a baseline for comparing the effects of distance in biometric performance. 

\textbf{Long Range Face}: This category consists of data captured using long range cameras on the ground that are configured for facial imagery at a resolution suitable, though perhaps not ideal, for recognition. Long range face recognition is an active research area and encompasses surveillance use cases requiring that cameras be positioned at significant distances from their intended subjects. We are particularly interested in any recognition improvements that may be observed when leveraging body and gait imagery.

\textbf{Long Range Body (Fusion)}: This area includes all ground sensors at long range that are poorly configured for biometric recognition. This group is a super set of the long range face mission area which also includes body and gait, highlighting the improvements achieved by fusing these additional modalities along with the face. 

\textbf{UAV}: Imagery captured from UAV platforms introduces unique challenges relating to distance, pitch angle, and platform size and weight. Consistent, accurate performance using this data would constitute a large step forward in the realm of biometric recognition and would support many surveillance and security use cases in which UAV platforms are already used or static camera equipment on the ground would be intractable. 

\textbf{Turbulence}: Atmospheric conditions can degrade outdoor image quality significantly and lead to poor recognition performance. Turbulence visibly distorts imagery, particularly at medium and long ranges, and may challenge any biometric application which relies on data captured outside from any nontrivial distance. This problem varies based on environmental conditions like climate, weather, and terrain.  

\textbf{Elevated Cameras}: This area focuses on close range surveillance cameras positioned to look down from rooftops or masts at pitch angles exceeding \(12^{\circ}\). In many instances, this is the most suitable way to place cameras due to space constraints or concern over damage to surveillance equipment. Extreme pitch angles are challenging since they do not capture the frontal or eye-level face imagery that typically produces the most accurate recognition. 

\textbf{Gait}: Gait is a relatively nascent biometric modality when compared to those that are more established and widely adopted like the face. However, advancing gait recognition performance will serve many applications not only on its own but in conjunction with modalities like the face and body. Even as an auxiliary signal, gait patterns may provide more accurate recognition capabilities in challenging uncontrolled imaging conditions. 

\textbf{Face Restricted}: This area targets data from long-range cameras, elevated cameras, and cameras mounted to UAV platforms. Faces are totally or partially occluded, observed from extreme pitch angles, or are of lower resolution than 20 pixels in height. Improving recognition in the absence of clear facial imagery would be advantageous for a wide range of use cases in which subjects are captured in an uncontrolled environment not suitable for close range, ground level equipment installation. While previous research has shown that incorporation of degraded facial imagery can improve identification performance~\cite{li2016toward}, recognition under these conditions still requires that algorithms not rely on facial imagery but instead make accurate predictions using modalities such as body and gait. 

No biometric modality or system is suitable for all applications. Stakeholders may care about one or two particular mission areas and will focus on results for those use cases. For example, many organizations are interested in recognition from cameras mounted on building rooftops while others may be better served by long-range systems. We selected the mission areas detailed above because they encompass many recognition applications and use cases for which there is need by government stakeholders.

\begin{table*}
\centering
\caption{Each category's combination of working distance/angle and biometric modality is reflective of a specific application of interest.}
\small
\begin{tabular}{lllll}
\toprule
\textbf{Mission} & \textbf{Description} & \textbf{Distance} & \textbf{Modality} & \textbf{\begin{tabular}{@{}l@{}}Subjects/\\ Samples \end{tabular}} \\
\midrule
Experimental Control&\textit{High- resolution surveillance cameras at eye-level}&3.8m-17.2m&Face, Body, Gait & 256/2282\\ 
\hline
Close Range Face&\textit{\begin{tabular}{@{}l@{}}High-resolution surveillance cameras on the ground or \\ slightly elevated, frontal and profile angles of the face \end{tabular}}&3.8m&Face& 245/1199\\ \\ \hline
Close Range Body  & \textit{\begin{tabular}{@{}l@{}}High-resolution surveillance cameras on the ground or  \\ slightly elevated  \end{tabular}} &3.8m  & Face, Body, Gait& 250/1719\\ \\ \hline
Long Range Face & \textit{Cameras configured for face recognition on the ground} & \(>\)250m & Face& 226/1351 \\
\hline
 \begin{tabular}{@{}l@{}}Long Range Body \\ (Fusion)\end{tabular} & \textit{\begin{tabular}{@{}l@{}} Cameras designed for long range but poorly configured \\ for biometric recognition \end{tabular}} & \(>\)250m  & Face, Body, Gait& 258/3166 \\ \hline
UAV & \textit{\begin{tabular}{@{}l@{}}Video from UAV platforms at varied elevation, pitch angle, \\ movement speed\end{tabular}} & - & Face, Body, Gait& 44/238 \\ \hline
Turbulence & \textit{Long range video reflecting medium to high turbulence} &  \(>\)250m  & Face, Body, Gait& 232/3272 \\ \hline
Elevated cameras & \textit{Close range cameras at severe pitch angles} & 5.8m-12.9m & Face, Body, Gait& 242/1532 \\ \hline
Gait & \textit{Video of walking sequences from all angles} & \(>\)3.8m  & Gait& 260/3744 \\ \hline
Face Restricted & \textit{Video lacking sufficient facial quality for face recognition} & \(>\)5.8m  & Body, Gait& 260/2078 \\ 
\bottomrule
\end{tabular}

\label{tab:mission}

\end{table*}
\section{Analysis Design}
\label{sec:analysis-design}
Evaluating biometric systems based on specific mission areas informs development of applications targeting use cases that fall within these areas. It is useful to consider factors that may improve performance in one domain while hindering it in another. We design experiments targeting these mission areas in greater isolation than would be captured by a more general performance evaluation. Additionally, this analysis design is based on the idea that, under challenging conditions, systems will perform better when they can leverage imagery of both the face and the body than facial imagery alone. To this end, these experiments measure the performance of the selected algorithms on face, body, and/or WB fusion (face, body, and gait) recognition tasks using a dataset that consists of two distinct probe sets, Face Included and Face Restricted. 
For the purposes of evaluation we use probe videos from field data and gallery enrollments from controlled data. Probe videos contain long-range or elevated views for a single subject. After processing a video, a labeled database entry for the subject is made. It is assumed that each entry could have multiple tracklets and each tracklet would be associated with a template matching an entry in the gallery. 
Gallery enrollments are a collection of templates associated with subjects. An enrollment produces a single template entity that can be searched against and matched. And the template encapsulates features extracted from one or more pieces of input imagery from a subject. The imagery could be face and/or WB images and videos.
Evaluations are implemented by creating probe and gallery databases. API commands are provided to verify against these databases. A ROC curve, which plots the true accept rate (TAR) vs the false accept rate (FAR) over the range of thresholds, is generated for each mission area. The experiments can be run in different modes such as face only, body only, gait only and fusion to highlight the strengths and weaknesses of systems as well as to show improvements, if any, when fusion is selected. As mentioned in Section ~\ref{sec:mission-areas}, the main focus of this evaluation analysis is to use mission areas to better utilize the data to analyze recognition performances with experimental control as the baseline for comparison. In addition, investigating recognition performance between two data treatments, such as close range and long range, highlights the effects of certain factors such as distance.

\section{Results and Discussion}
\label{sec:results-and-discussion}



\begin{figure}
    \centering
    \includegraphics[width=0.95\linewidth]{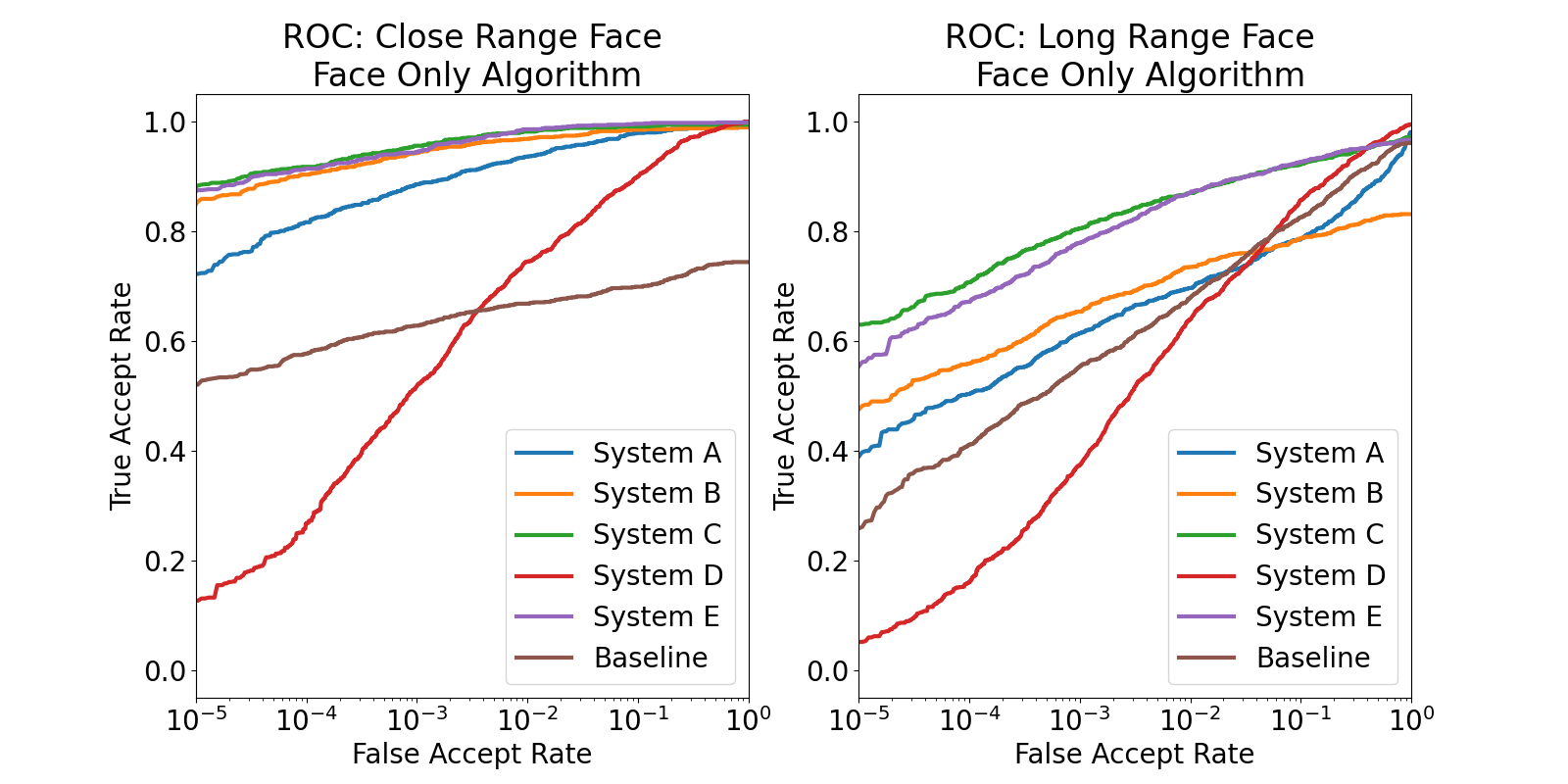}
    \caption{Close range, long range - face performance results}
    \label{fig:close-long-range-face}
\end{figure}

\textbf{Face Recognition}: The Close Range Face mission serves as a basis for comparison and consists of high-resolution imagery representative of high-quality video from current surveillance camera deployments. Face recognition algorithms should be able to perform well with existing technology. The Long Range Face mission presents additional challenges related to image quality, allowing us to test both the basic functionality of the algorithms and understand how new developments are improving accuracy in more challenging scenarios. ROC results are compared in Figure \ref{fig:close-long-range-face}, which shows promising performance.

\begin{figure*}
    \centering
    \includegraphics[width=0.65\linewidth]{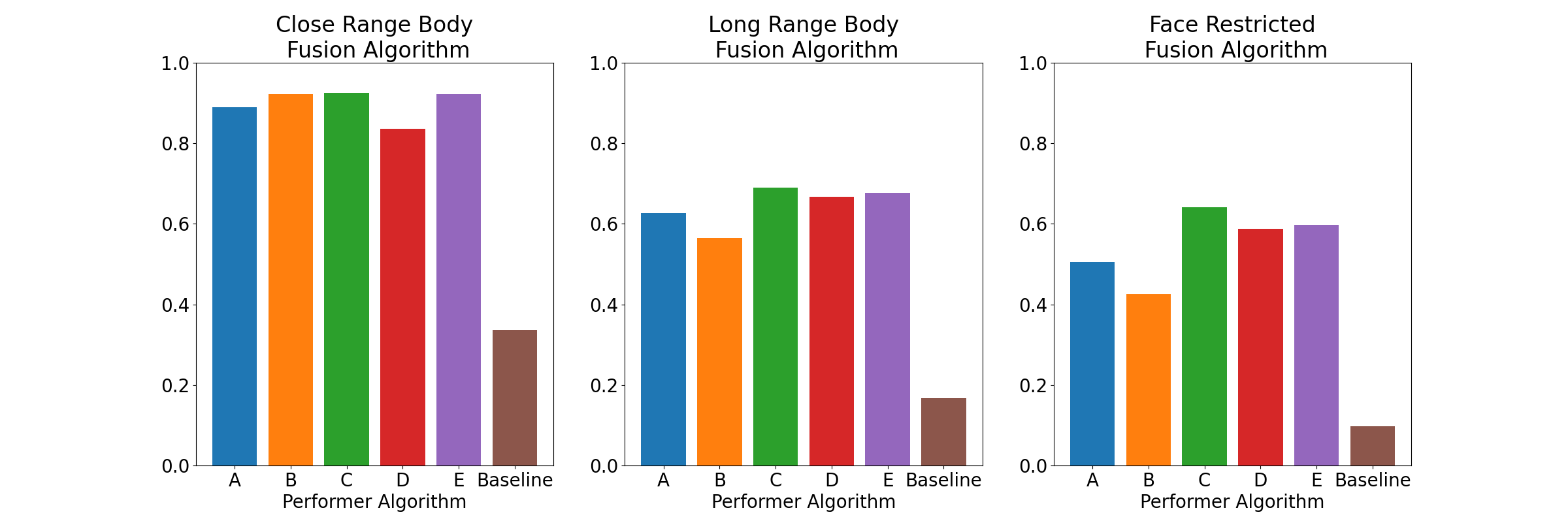}
    \caption{Close range body, long range body, face restricted - fusion results compared for all systems with the performance metric of TAR @1$\%$ FAR}
    \label{fig:long-range-body-all}
\end{figure*}

\textbf{Whole Body Matching}: One of the primary goals of this research is to develop techniques for recognizing individuals based on their whole-body appearance. Combining face recognition with body geometry, gait, and other features should improve matching performance. This is particularly important given the image quality challenges in our dataset. Figure \ref{fig:long-range-body-all} compares results in this area and shows how performance decreases under different levels of difficulty. Close range shows very high accuracy, while longer ranges show systems performing well despite being poorly configured for recognition. Face restricted performance is surprisingly good given that these videos were selected in a way that well-established face recognition algorithms lacked usable faces, requiring the algorithms to rely heavily on body and gait features.  

\begin{figure*}
    \centering
    \includegraphics[width=1.0\linewidth]{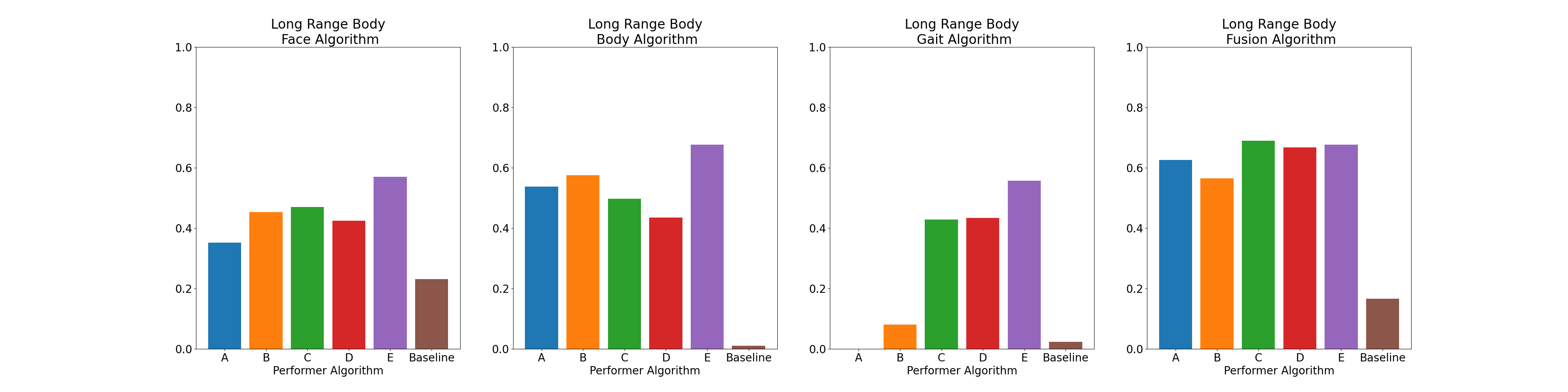}
    \caption{Results from the system with individual modalities isolated vs the fusion.  All algorithms are run on the long range body mission. The performance metric compared is TAR @1$\%$ FAR}
    \label{fig:long-range-fuse}
\end{figure*}

\textbf{Improvements from Fusion}: Another critical area of interest is how the fusion of different biometric modalities improves recognition accuracy. Figure \ref{fig:long-range-fuse} illustrates how face only, body only, and gait only recognition is combined to produce improved fusion accuracy. The body only case outperforms the face only case for most of the systems and fusion of all modalities help the overall recognition performance. Based on this analysis, the challenges that come from identifying individuals at long range would be greatly reduced when all modalities are utilized. By analyzing individual modalities as well as fused modalities across a variety of missions, we can better understand the contributions of each modality as well as the value of fusion.

\textbf{Elevated and UAV}: At this early stage of the program, elevated and UAV performance have not yet been a key focus. However, providing tests in these areas early on is intended to guide future research in this field.

\begin{figure}
    \centering
    \includegraphics[height=1.25in]{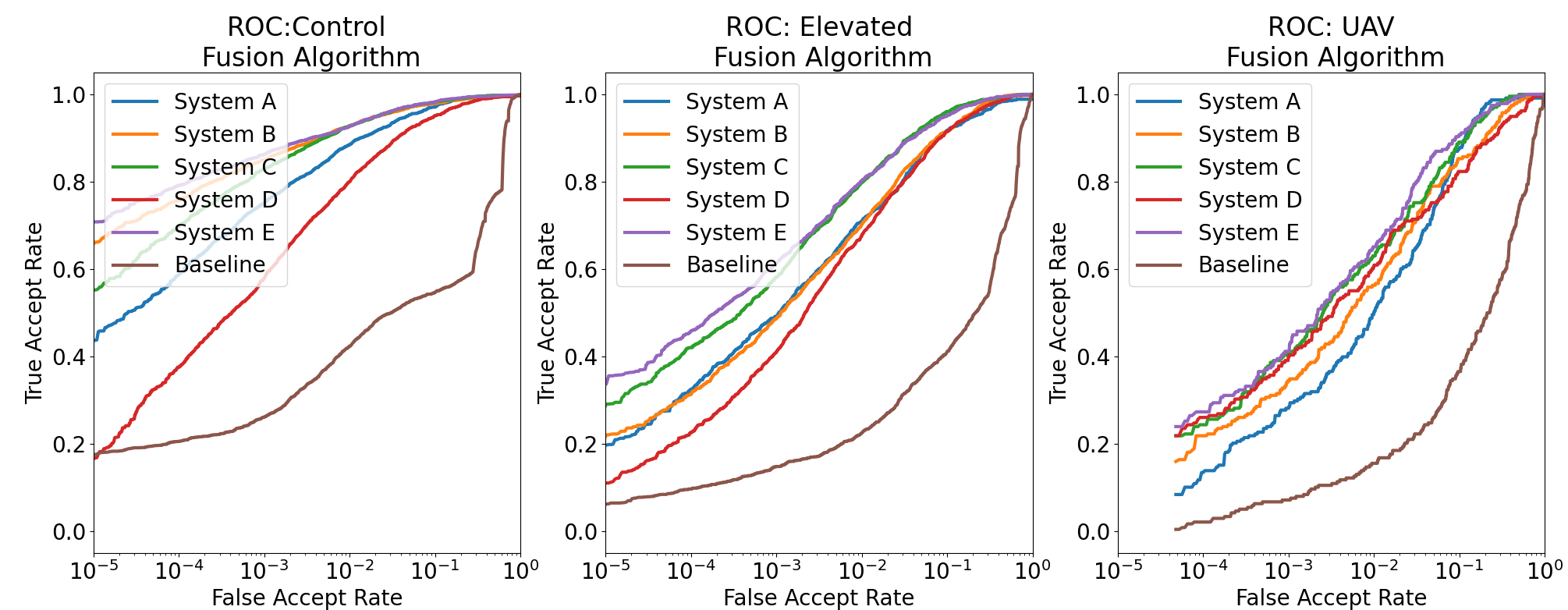}
    
    \caption{Control, Elevated, UAV - fusion performance results}
    \label{fig:elevated-fuse}
\end{figure}

Current results offer a snapshot of performance, showing that developed algorithms significantly outperform the baseline, which has not been updated to handle high pitch angle data. Comparing the Experimental Control (Close-range Ground) test to the Elevated Mission (Close-range Elevated) illustrates the challenges posed by these scenarios as seen in Figure \ref{fig:elevated-fuse}. The UAV mission is currently intended to focus development on some of the most difficult challenges in this research program. We anticipate seeing significant improvements in these areas in the next year of research.

\begin{figure}
    \centering
    \includegraphics[height=1.25in]{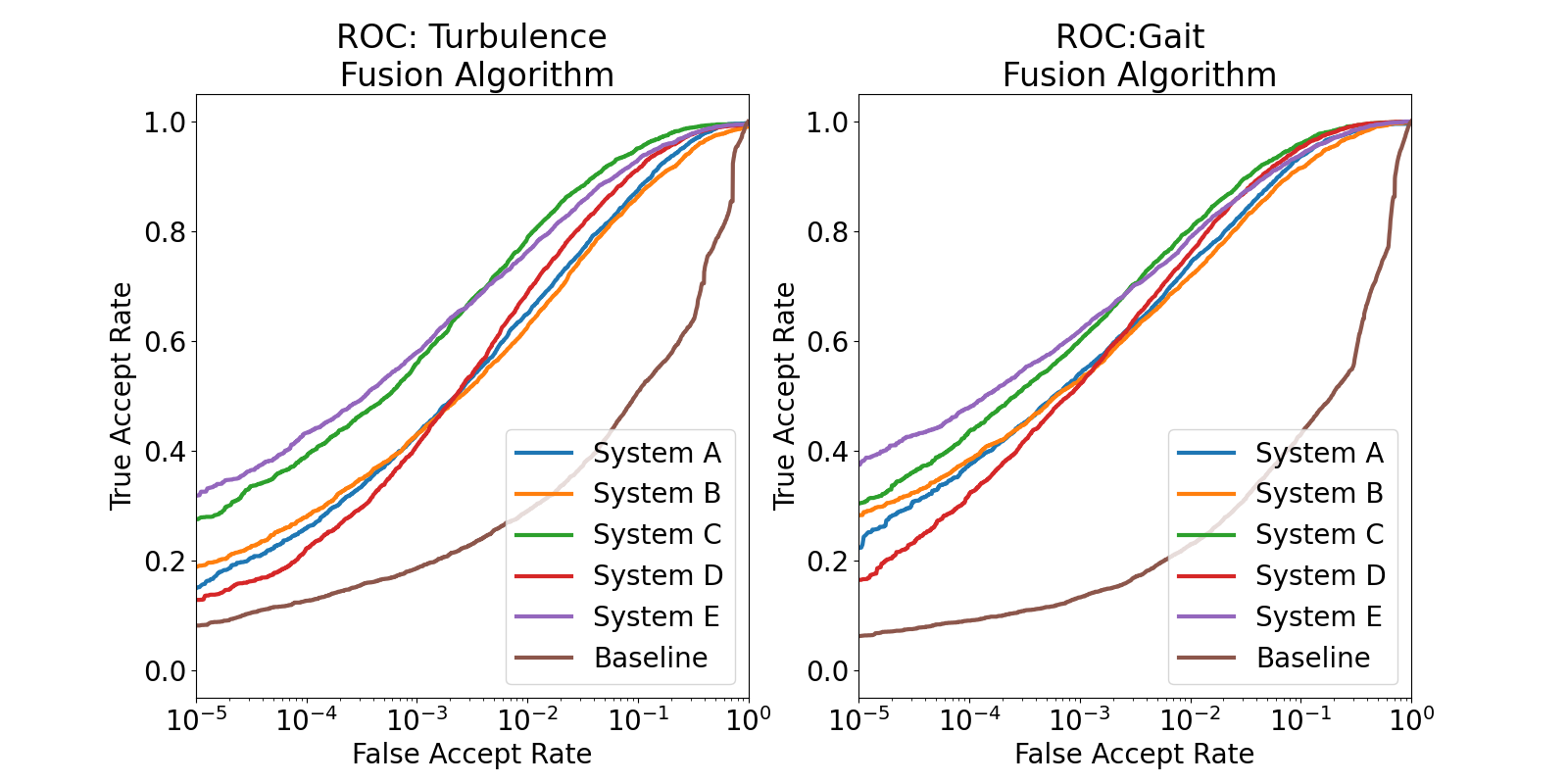}
    \caption{Turbulence, Gait - fusion performance results}
    \label{fig:turb-gait}
\end{figure}

\textbf{Turbulence and Gait}: The turbulence and gait missions are used to focus algorithm development on two specific problems: gait recognition and turbulence mitigation. The gait mission provides a test set where all media contains people walking, providing a good way to test gait recognition in isolation on a challenging dataset. Likewise, the turbulence mission focuses on data known to be at long distances and with significant amounts of turbulence. By splitting these problems out, it allows us to focus research and development in these specific areas, benchmark progress over time, and compare different solutions. Results are shown in the ROC plots in Figure \ref{fig:turb-gait} which demonstrate current performance on these two problems. It should be noted that the Baseline algorithm uses Body and Gait recognition algorithms that predate the data collection and are known to be inadequate for the level of difficulty of this data set. System A also did not include a gait algorithm, but provided body features that still perform competitively with its peers.

\section{Conclusion and Future Work}
The design of thoughtful experiments that help move the needle on SOTA biometric algorithms requires a tight marriage between physical data collection processes and the detailed curation and partitioning of imagery and video therein. We hope this paper acts as a thought-provoking case study into that process, along with providing a rich analysis of anonymous SOTA algorithms tested against our experimental protocols.

Our evaluation methods addresses challenges related to image quality and focuses on reducing the dependence on face recognition as the sole modality by providing data and tests focused on whole body matching. The evaluation utilized a diverse range of hardware solutions including COTS cameras, custom long-range cameras, and UAVs. Preliminary results indicate promising progress in WB recognition. 

In future work, we intend to further address challenges related to image quality and dependence on face recognition as the primary modality in long-range biometric identification systems. We aim to promote development of robust systems that are less susceptible to these drawbacks. We also plan to focus on video recordings featuring larger groups of individuals as well as more intricate and dynamic scenarios. By doing so, we hope to continue encouraging development of more accurate and reliable methodologies for WB recognition technology. We are confident that these advanced systems will have significant implications for various mission areas, including law enforcement and national security. Further leveraging WB recognition capabilities will potentially improve situational awareness, public safety, and operational efficiency.


\section*{Acknowledgements}

This research is based upon work supported by the Office of the Director of National Intelligence (ODNI), Intelligence Advanced Research Projects Activity (IARPA), via D20202007300010. The views and conclusions contained herein are those of the authors and should not be interpreted as necessarily representing the official policies, either expressed orimplied, of ODNI, IARPA, or the U.S. Government. The U.S. Government is authorized to reproduce and distribute reprints for governmental purposes notwithstanding any copyright annotation therein.

This research used resources from the Knowledge Discovery Infrastructure at the Oak Ridge National Laboratory, which is supported by the Office of Science of the U.S. Department of Energy under Contract No. DE-AC05-00OR22725

Notice:  This manuscript has been authored by UT-Battelle, LLC, under contract DE-AC05-00OR22725 with the US Department of Energy (DOE). The US government retains and the publisher, by accepting the article for publication, acknowledges that the US government retains a nonexclusive, paid-up, irrevocable, worldwide license to publish or reproduce the published form of this manuscript, or allow others to do so, for US government purposes. DOE will provide public access to these results of federally sponsored research in accordance with the DOE Public Access Plan
(\url{http://energy.gov/downloads/doe-public-access-plan}).

\vspace{20px}

{\footnotesize
\bibliographystyle{ieee}
\bibliography{egbib}
}

\end{document}